# Enhancing human bodies with extra robotic arms and fingers: The Neural Resource Allocation Problem

Giulia Dominijanni[1,2 *], Solaiman Shokur[1,2 *], Gionata Salvietti[3,4], Sarah Buehler[5], Erica Palmerini[6], Simone Rossi[7], Frederique De Vignemont[8], Andrea D'Avella[9,10], Tamar R. Makin[5 ‡], Domenico Prattichizzo[3,4 ‡], Silvestro Micera[1,2 ‡ □]

1. Bertarelli Foundation Chair in Translational NeuroEngineering, Center for Neuroprosthetics and School of Engineering, Ecole Polytechnique Federale de Lausanne (EPFL), Lausanne, Switzerland

2. Institute of BioRobotics and Department of Excellence in Robotics and AI, Scuola Superiore Sant'Anna, Pisa, Italy

3. Department of Information Engineering and Mathematics, University of Siena, Siena, Italy

4. Department of Advanced Robotics, Istituto Italiano di Tecnologia, Genoa, Italy

5. Institute of Cognitive Neuroscience, University College London, UK

6. Dirpolis Institute, Scuola Superiore Sant'Anna, Pisa, Italy

7. Siena Brain Investigation & Neuromodulation Lab (Si-BIN Lab), Department of Medicine, Surgery and Neuroscience, Unit of Neurology and Clinical Neurophysiology, University of Siena, Italy

8. Institut Jean Nicod, CNRS-ENS-EHESS, PSL University, France

9. Department of Biomedical and Dental Sciences and Morphofunctional Imaging, University of Messina, Italy

10. Laboratory of Neuromotor Physiology, IRCCS Fondazione Santa Lucia, Rome, Italy

[*] These authors contributed equally to this work

[‡] These authors contributed equally to this work

[□] Corresponding author

## Abstract

The emergence of robot-based body augmentation promises exciting innovations that will inform robotics, human-machine interaction, and wearable electronics. Even though augmentative devices like extra robotic arms and fingers in many ways build on restorative technologies, they introduce unique challenges for bidirectional human-machine collaboration. Can humans adapt and learn to operate a new limb collaboratively with their biological limbs without sacrificing their physical abilities? To successfully achieve robotic body augmentation, we need to ensure that by giving a person an additional (artificial) limb, we are not in fact trading off an existing (biological) one.

In this manuscript, we introduce the *"Neural Resource Allocation"* problem, which distinguishes body augmentation from existing robotics paradigms such as teleoperation and prosthetics. We discuss how to allow the effective and effortless voluntary control of augmentative devices without compromising the voluntary control of the biological body. In reviewing the relevant literature on extra robotic fingers and limbs we critically assess the range of potential solutions available for the *"Neural Resource Allocation"* problem. For this purpose, we combine multiple perspectives from engineering and neuroscience with considerations from human-machine interaction, sensory-motor integration, ethics and law. Altogether we aim to define common foundations and operating principles for the successful implementation of motor augmentation.

## I. Introducing robotic body augmentation and the neural resource allocation problem

With robotic body augmentation we are witnessing the rise of a new class of robotic technologies, which are designed to resemble human limbs in their functionality while being integrated with the users' physical abilities. Traditionally, such devices have been developed to substitute a missing or impaired body function, most famously bionic legs and arms for substitution of missing limbs *(1, 2)* or exoskeletons for restoring impaired movement *(3)*. But in principle, from a system design perspective, once a functionality that approximately matches that of a body part can be implemented, the same technological foundation can be exploited for augmenting the sensory and motor capabilities of an able-bodied individual using extra robotic limbs. As such human body augmentation is no longer science fiction. From the engineering side, a whole spectrum of human enhancement now exists, ranging from technologies for restoration or compensation of functions in patients with physical limitations (Fig. 1A-B) to augmentation beyond a (healthy or disabled) subjects' physical abilities (Fig. 1C-D) *(4)*.

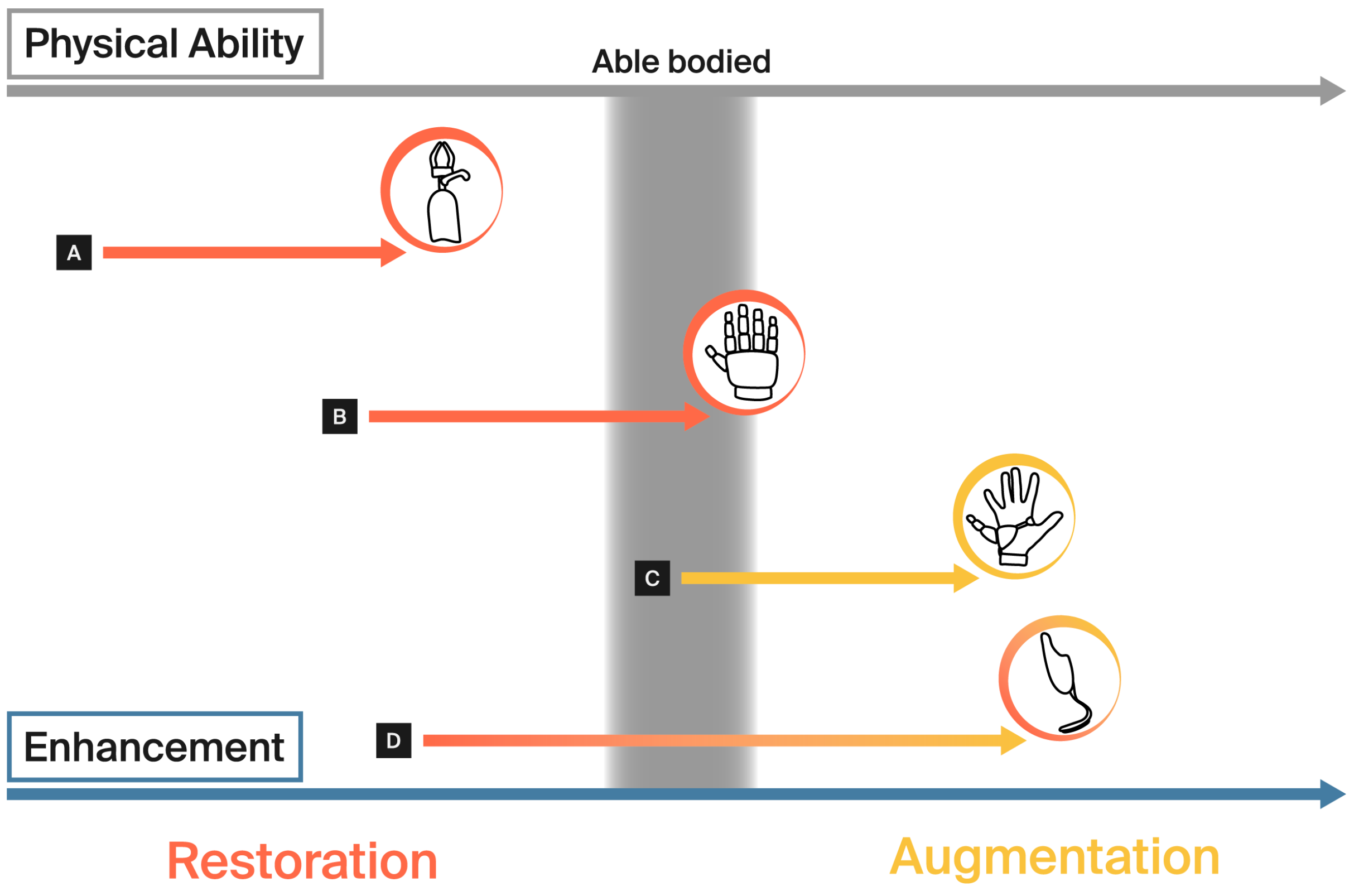

**Figure 1:** Enhancement continuum and examples of restoration (orange) and augmentation (yellow) of a function (e.g., reaching and grasping) with respect to bodily ability. The partial (**A**) or complete (**B**) restoration of a function in a subject with a given initial level of impairment. (**C**) The enhancement of able-bodied subjects for a given function is defined as augmentation. (**D**) The enhancement of a subject with disabilities beyond the capabilities of an able body is also a case of augmentation.

The examples in Fig. 2A show use cases of sensorimotor augmentation that are achievable with the current state of the art technology, such as extra robotic arms that enable holding and screwing simultaneously or extra robotic fingers that stabilize a grip while opening a jar with only one hand. But more complex applications are likely to emerge in the future (Fig. 2B). Recent achievements in bidirectional human-machine interfaces pave the way for future "augmented bodies," introducing the possibility that restorative and augmentative technology might eventually become two sides of the same coin *(5)*.

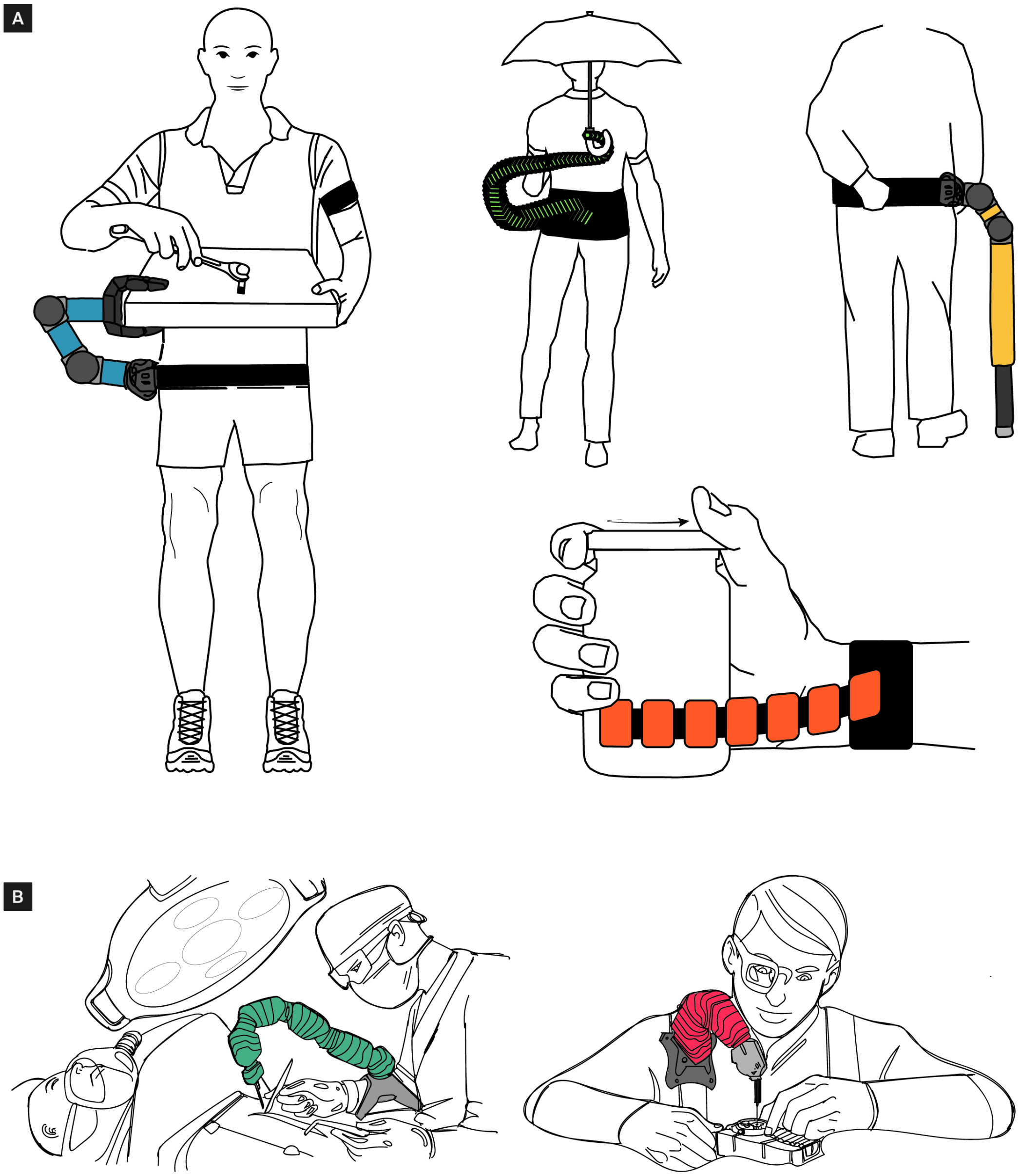

**Figure 2:** (**A**) Examples of extra robotic limbs use scenarios on the basis of the current state of the art in the field. In blue, an extra robotic arm helps in holding and screwing simultaneously; in green, a robotic tail holds an umbrella freeing the wearer's hands; in yellow, an extra leg helps an elderly walking; in orange, an extra robotic finger helps stabilize the grip while opening a jar with only one hand. (**B**) Future possible use scenarios of extra limbs in the medical and industrial applications. In green, a doctor performs a surgery with the help of an extra limb; In red, a watchmaker screws a component with an extra robotic limb while holding the watch with his natural hands.

Extra robotic arms (XRAs) and extra robotic fingers (XRFs) are unique devices that share attributes with existing robotics paradigms — such as prosthetics, wearable robotics, teleoperation, and human-robot collaboration — and yet possess unique features and functionalities. This warrants augmentation a new field with its own scientific and technological foundations. Conceptually, XRFs and XRAs sit in an unexplored region of the threedimensional space defined by the control strategy with respect to the biological limbs, level of enhancement, and wearability (Fig. 3A). Under the augmentation framework, they are a wearable technology that can be controlled independently and/or simultaneously with the biological limbs. In terms of enhancement, they augment (i.e., add to) a users' physical abilities (Fig. 3B), rather than substituting a lost function (e.g., Fig. 3C, (v) exoskeletons or (vi) prosthetics), or rerouting an existing function *(6)* (e.g., Fig. 3C, (vii) robotic arms used in teleoperation). Moreover, they do not rely on an autonomous agent that interprets human intentions (e.g., (viii) collaborative robotics) but instead are controlled at the user's own will. As such, while some technical challenges for augmentation and substitution technologies are similar (e.g. developing light-weight wearable systems or increasing battery duration), robotic augmentation control provides new conceptual and practical challenges, which we would like to explore here.

As an extension of the user's sensorimotor system, XRAs and XRFs are meant to be simultaneously controlled with the biological arms and fingers in an effective and intuitive way. This integration with the human body poses the urgent question of whether and how the human brain can support the control of extra-limbs. People born with six fingers, for example, have dedicated nerves, muscles, and sensorimotor cortical representations to control the extra finger and achieve good motor performance *(7)*. Humans wearing extra robotic limbs cannot count on such a dedicated "neural hardware," so they need to adjust their "software" (i.e., neural activities) to efficiently control the extra robotic limb *(6)*. Existing solutions involve 'highjacking' the neural resources originally devoted to our own body. But for augmentation, much unlike substitution, we cannot harness "freed up" resources, such as the residual nerves in the arm of an amputee used for controlling a prosthesis. Moreover, unlike non-wearable teleoperation, augmentation should not interfere with the motor control of a user's biological limbs. Therefore, the challenge lies in operating the robotic limb without incurring costs to the rest of the body.

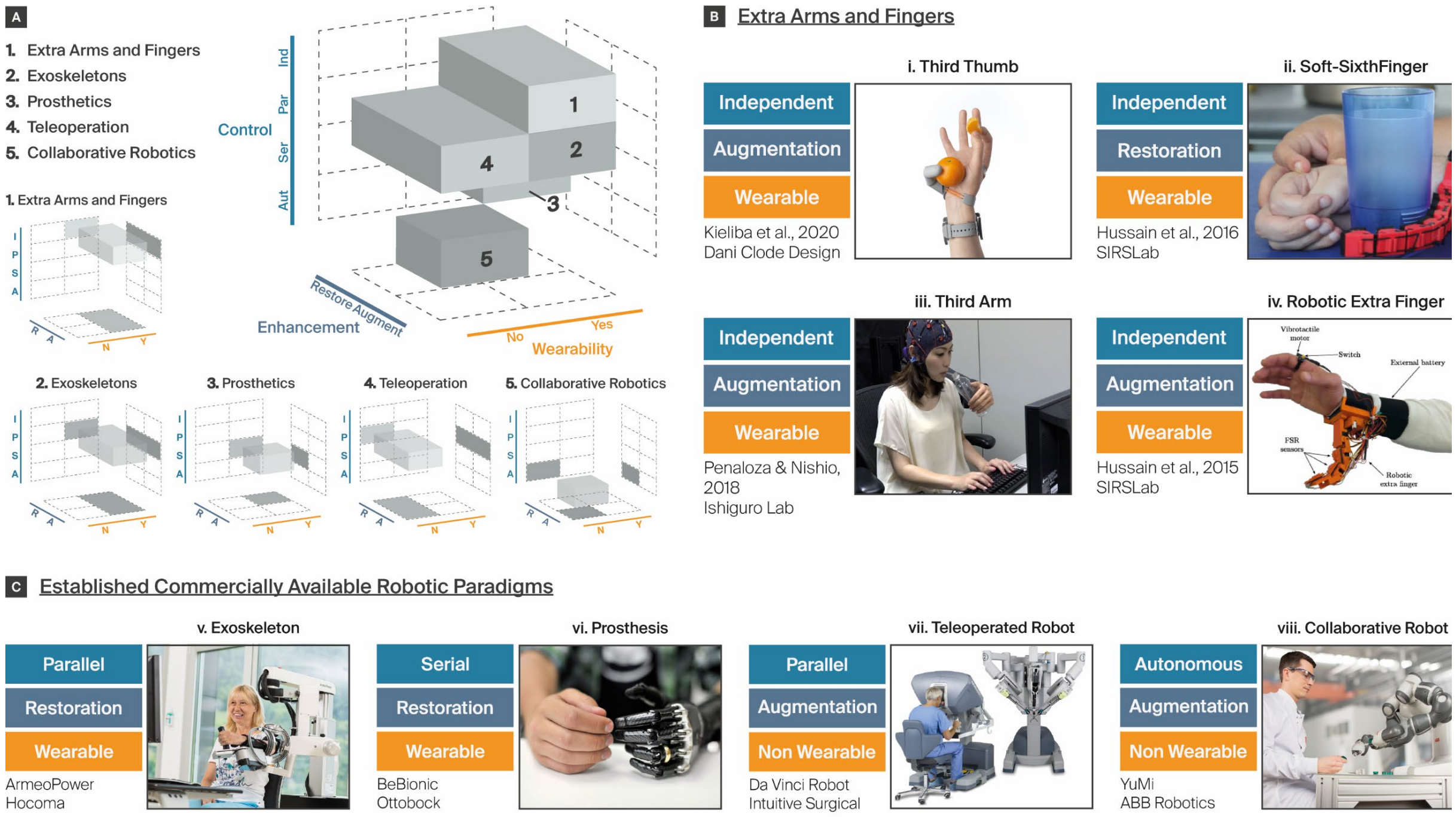

**Figure 3:** (**A**) Three-dimensional space defined by control strategy, enhancement, and wearability. **Control** is defined with respect to the natural upper limbs' joints: robotic devices can be controlled (Ind)ependently from the user's upper limbs, in (Par)allel by mirroring the user's biological limbs, in (Ser)ies by extending the kinematic chain of the biological limbs or they could be entirely (Aut)onomous. The **Enhancement** is divided into two categories: restoration (Restore) and augmentation (Augment); **Wearability** defines whether the device is meant to be worn (Yes) or not (No);

(**B**) Examples of XRAs/XRFs with their corresponding description for control (light green), enhancement (dark green), wearability (orange). (i) The third thumb, designed by Dani Clode is controlled using force sensors strapped underneath the participant's big toes *(13)*. (ii) The soft Sixth finger is an extra finger used for the restoration of functionality in stroke patients *(11)*; the finger is controlled via decoding of muscle activation. (iii) A healthy subject can control a human-like robotic arm for multitasking, using an EEG-based brain-machine interface *(9)*. (iv) The robotic extra finger is controlled via a switch on a ring; the same ring provides vibrotactile feedback to the user *(10)*.

(**C**) Representative examples of classes of established robotic paradigms. (v) The ArmeoPower Upper-limb exoskeleton, for hand therapy (Hocoma AG, Volketswil, Switzerland). (vi) The bebionic prosthetic hand for transradial amputee patients (Ottobock, Leeds, England). (vii) The da Vinci robot is a tele-operated surgical robot (Intuitive Surgical, Sunnyvale, USA). (viii) YuMi robot permits safe collaboration between the user and the robot (ABB, Zurich, Switzerland).

Hereafter, we will refer to this unique challenge, which distinguishes augmentation technology from most other assistive and restorative technologies, as the "*Neural Resource Allocation"* problem. Put simply, this denotes the channelling of motor commands and sensory information to and from the augmentative device without hindering the motor control of biological limbs.

In the following, we wish to critically assess the range of potential solutions available to address the "*Neural Resource Allocation"* problem when designing bidirectional control strategies for a new body part, from both a neuroscience and engineering perspective. We will review the first technologies pioneering the field, specifically focusing on XRFs and XRAs. While these technologies provide a first proof-of-concept for the feasibility of upper limb motor augmentation, they also highlight the many technical and conceptual challenges that are still unresolved. As such we will consider how to best allow the effective and effortless voluntary control of these devices without compromising the voluntary control of the biological limbs. Further, we will address the problem of providing somatosensory information about the state of these devices without interfering with the sensory inputs coming from the biological limbs. Finally, we will highlight some of the societal, ethical, and legal aspects of extra limbs, which cannot be excluded from a discussion on the future of this emerging technology. Together, we aim to address the "*Neural Resource Allocation"* problem and define common foundations and operating principles for the successful implementation of motor augmentation.

## II. Motor control of extra robotic arms and fingers

As highlighted above, the operation of XRFs and XRAs requires the coordinated motor control of a robotic limb without the physiological infrastructure to guide these movements. Therefore, to operate the device, motor resources devoted to another (biological) body part need to be employed. Hereby, the key challenge is to provide a reliable readout of motor signals, while minimally disrupting the functionality of the biological body part. Several approaches have been proposed, exploiting – among others – muscle or brain interfaces to achieve coordination with the biological limbs *(8, 9)*.

| **Reference** | **Type of extra limb** | **Control strategy** | **Sensory feedback approach** | **Number of subjects** | **Test task** | **Results** | **Classification** | | |
|---|---|---|---|---|---|---|---|---|---|
| | | | | | | | C | E | W |
| **Salvietti et al., 2017** *(8)* | Robotic extra finger for restoration | Forehead frontalis muscle sEMG interface. High level control strategy based on #contractions | None | 4 chronic stroke patients | Frenchay arm test (measure of upper extremity performance in daily activities) | All the subject increased their score from 1/5 to 3/5 | Independent, task-extrinsic muscular null space | R | Y |
| **Penaloza and Nishio, 2018** *(9)* | Robotic extra arm for augmentation | Single channel PSD threshold-based BMI. Relevant channel chosen during calibration (separately for single and multitask) | None | 15 healthy subjects | Single task bottle grasping/releasing and multitask which adds a ball balancing task in parallel | Percentage of correct trial is unimodal for single task (67.5% median) and bimodal for multitask (72.5% median). Good performers and bad performers had 85% and 52.5% median correct trial percentage respectively | Independent, neural null space | A | Y |

| | | | | | | | | |
|---|---|---|---|---|---|---|---|---|
| **Hussain et al., 2015** *(10)* | Robotic extra finger for augmentation | Switch button on a ring | Different vibrotactile feedback on a natural finger for encoding contact onset/offset or the exerted force | 10 healthy subjects | Pick-and-place task (20 randomized trials, 5 repetitions for each feedback condition + no feedback) | Haptic feedback improved task performances (completion time, exerted force, perceived effectiveness) and was preferred by the subjects with respect to no feedback. | Independent, task-intrinsic kinematic null space | A | Y |
| **Hussain et al., 2016** *(11)* | Robotic extra finger for restoration | EMG interface embedded in a cap (eCap) | None | 6 stroke patients | Frenchay arm test (measure of upper extremity performance in daily activities) and 4 bimanual tasks | All the subject increased their score in the Frenchay Arm test of at least 2/5 points. All managed to accomplish the bimanual tasks without training. | Independent, task-extrinsic muscular null space | R | Y |
| **Parietti et al., 2017** *(12)* | Robotic extra legs for augmentation | Sensor suit measuring sEMG from pectoral and abdominal muscles | None | 8 healthy subjects | Tracking task: control the extra legs, either in a simulation (exp 1) or wearing the prototype (exp 2) following two targets displayed on screen | The velocity control was identified as the best strategy (compared to position and muscle model) to control the simulated extra legs Both Naive and trained subjects performed better (accuracy and learning speed) when wearing the physical prototype (compared to average performance with the simulation) | Independent, task-extrinsic muscular null space, | A | Y |
| **Kieliba et al., 2020** *(13)* | Robotic extra finger for augmentation | Force sensors under the big toes (1 for flex/ext, 1 for abd/add) | None | 31 healthy subjects (20 augmentation, 11 control group) | A battery of behavioural tasks, looking at hand-XF coordination and collaboration, hand motor control (kinematics and force enslavement), body image, subjective sense of embodiment of the XF, hand representation using representational similarity fMRI analysis<br><br>MRI recording pre and post a 5-day training session with 'in the wild' usage | -Improved motor control, dexterity and hand-robot coordination (also in high cognitive load and occluded vision conditions) with training<br>-Increased sense of embodiment<br>-Modified kinematic hand synergies<br>- Modified hand motor representation | Independent, task-extrinsic kinematic null space | A | Y |
| **Saraiji et al., 2018** *(14)* | Robotic extra arms for augmentation | Wearable interface tracking the foot position/rotation and the toe posture | Force feedback to the sole employing a motor driven belt mechanism | 12 healthy subjects | One arm pointing task (only with the robotic arm) | Mean throughput of 1.01bit/s<br>No learning over the sessions<br>Agency but not ownership | Independent, task-extrinsic kinematic null space | A | Y |
| **Abdi et al., 2015** *(15)* | Virtual extra hand for augmentation | Foot movement tracking with depth cameras (MS Kinect) | None | 13 healthy subjects | 3 games: 1 hand target reaching, 3 hands sliding (third hand opposite to naturals), 3 hands falling objects catching | Use of the 3 virtual hands with no special priority<br>Fast learning | Independent, task-extrinsic kinematic null space | A | |
| **Wu and Asada, 2016** *(16)* | Robotic extra fingers | Bio-artificial grasp synergies | None | N.A. | Grasping of large objects success rate | The subject can learn to adapt his hand postures to avoid grasping failure (miss, slippage, extra fingers in the way) with the 7-fingered hand | Autonomous | A | Y |
| **Ciullo et al., 2020** *(17)* | Robotic extra hand for restoration | 3 input methods tested:<br>residual grasp force; bending sensor interface on the unaffected hand; trigger controlled by the unaffected hand. | None | 10 chronic stroke patients (FMA < 2) | Modified ARAT tasks | Score >13/30 for each patient with at least one interface (excl. dropouts) | Independent, task-extrinsic kinematic or muscle null space | R | Y |
| **Aoyama et al., 2019** *(18)* | Robotic extra finger for augmentation | Posterior auricular muscle sEMG interface | Vibrotactile phantom sensation interface on the contralateral hand | 1 healthy subject | Timed extra finger opposition task with and without feedback | Learning effect: # successful hits per trial increases throughout the experiment<br>Feedback reduces the average unsuccessful hits ($p < 0.1$) | Independent, task-extrinsic, muscular null space | A | Y |
| **Guggenheim et al., 2020** *(19)* | Robotic extra limbs for augmentation | Torso muscles sEMG interface | None | 11 healthy subjects | Pointing task: minimize position error between targets and limbs (natural and/or robotic) | Performances with natural limbs worsen when adding extra limbs (though the opposite is not true)<br>The four limbs are seldom moved concurrently, instead simultaneous movement of the natural followed by simultaneous movement of the robotic limbs was reported | Independent, task-extrinsic muscular null space | A | Y |

| | | | | | | | | |
|---|---|---|---|---|---|---|---|---|
| **Guggenheim et al., 2020** *(20)* | Robotic extra arm for augmentation | Patterns of fingers pressure | None | N.A. | Door opening and walk through | Successful demo | Independent, task-intrinsic kinematic null space | A | Y |
| **Lisini Baldi et al., 2020** *(21)* | Robotic extra finger for augmentation | Dominant arm kinematic recordings | None | 10 Healthy subjects | 2 tasks in virtual reality (trajectory tracking and spheres overlapping) and 2 tasks in real environment (single and multiple objects pick and place) | The 2 experiments in virtual reality proved the control strategy to be appropriate when used alone and to control an extra degree of freedom. The real environment experiments demonstrated that subjects can learn to use the system to perform pick and place tasks. | Independent, task-intrinsic kinematic null space | A | Y |

**Table 1**: Summary of previous studies on XRA and XRFs sensorimotor control and assessment, further categorized with respect to the control (C), enhancement (E), wearability (W) space defined in Figure 3. The two categories of enhancement are reported in the table as R (restoration) and A (augmentation). Y (yes) and N(no) describe whether the device is meant to be worn.

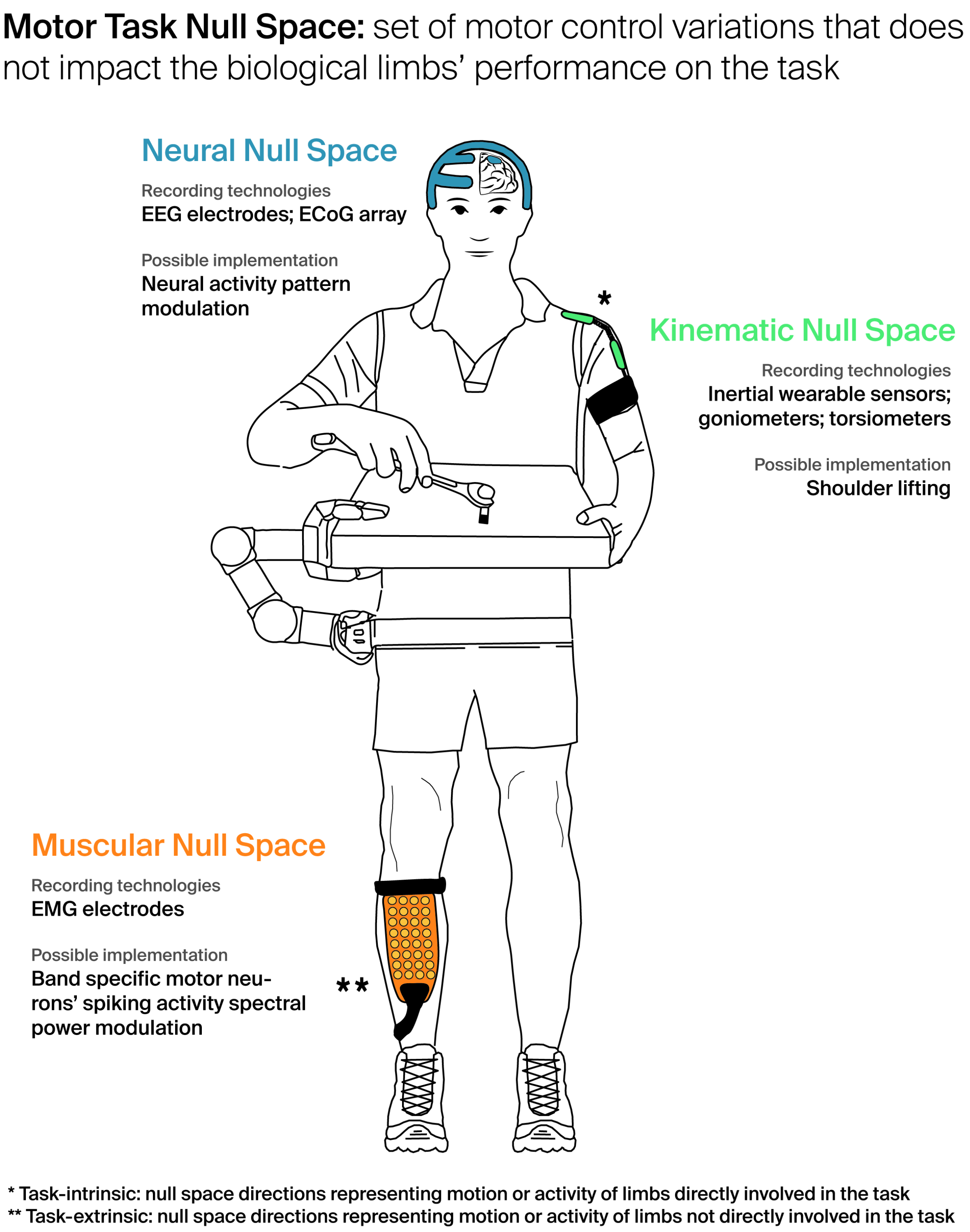

**Figure 4:** Examples of the kinematic (green), muscular (yellow) and cortical (purple) motor task null spaces with their recording technologies and possible implementations. One astericks denotes task-intrisic null space implementation while two asterisks task-intrinsic ones.

In light of this, an important concept we want to introduce here is the "**motor task null space**" (Fig. 4). To define the null space, we need to refer to a motor task that requires movement of the biological limbs. Because of musculoskeletal and neural redundancy *(22)*, different but equivalent body motions, muscle activation patterns, and neural activity patterns can be used to perform the task. The motor task null space then geometrically describes the set of all motor control variations — at the kinematic, muscular, or neural level — that do not impact the biological limbs' performance on that task (see Table 1 for relevant examples from the literature). In other words, by exploiting motor control variations whose effect is negligible on the control of the biological limbs involved in the task, humans can simultaneously control extra robotic limbs. The motor task null space is at the very foundation of the enhancement. It is captured by the sensing part of the human-XRF/XRA interface and transformed into motor control commands for the device. The nature of the signal captured determines the type of motor task null space: (i) kinematic, (ii) muscular, and (iii) neural (Fig. 4).

The "**kinematic null space**" (Fig. 4) refers to a subset of vectors whose components are samples of kinematic variables (e.g., joint angles) that characterize the movements of the limbs and are captured by motion capture systems, akin to those based on wearable technology *(21, 23)*. As illustrated in Fig. 4, we can subdivide the null space into two orthogonal subspaces according to the kinematic variables involved. The "task-extrinsic null space" spans null space directions representing motions of all biological body parts not directly involved in the motor task. For instance, when force sensors are placed underneath a user's big toes to control two movement degrees of freedom, flexion/extension and adduction/abduction, of a robotic finger mounted on a healthy subject's hand (example (i) in Fig. 3B). In contrast, the "task-intrinsic null space" spans null space directions representing motions that are restricted to the limbs directly involved in the task. For example, in *(10)* a push button was used to control an XRF attached to the wrist of the same hand used to press the button (example (iv) in Fig. 3B). It is worth noting that even though the task-extrinsic null space might intuitively seem like a better choice because it does not inherently interfere with the task, it is not always a practical solution for real-world tasks that require all limbs (e.g., engaging in bimanual manipulation while walking). From a neuroscience perspective, the task-intrinsic null space might then offer more relevant resources.

The **"muscular null space"** (Fig. 4) refers to a subset of vectors whose components are samples of the variables that express the level of activation of a set of muscles (or motor units), as estimated from electromyography (EMG) signals captured by wearable electrodes. For instance, any muscular co-contraction that does not generate net joint torques is associated with a vector in the muscular null space. While it may overlap with the kinematic null space, since there are more muscles than joints, the dimensionality of the muscular null space is generally larger than that of the kinematic null space. As such, the muscular null space offers additional opportunities to interface with the motor system without requiring explicit movements. For example, the volitional modulation of the motor neurons beta-band activity *(24)*, to some extent independently from muscle contraction, belongs to the muscular null space. An advantage of the task-extrinsic muscular null space for an upper limb task is that it can elicit contraction of muscles with no action on the upper limb joints. This may involve using abdomen or forehead muscles for instance – both of which have previously been exploited for the control of XRAs and XRFs respectively (see example (ii) on Fig. 3B) *(8, 18)*. Meanwhile, the task-intrinsic muscular null space would involve the co-contraction of pairs of antagonist muscles (e.g., biceps and triceps) of the arm involved in the task. This could be exploited to control extra robotic limbs without interfering directly with the user's manipulation capabilities. However, while co-contraction can be modulated voluntarily *(25)*, it is used naturally to regulate the mechanical impedance of the arm rather than to control extra robotic limbs. Thus, task-intrinsic muscular null space control constitutes a new motor skill that requires learning and practice. It remains to be understood how fast users can learn such motor skills *(26)* and what performance can be achieved, e.g., how many extra degrees of freedom can be controlled simultaneously with the biological limb and how accurate is the resulting motor control *(27)*.

Finally, the **"neural null space"** (Fig. 4) refers to a subset of vectors whose components are neural activity signals that can be independently modulated while performing the motor task. These can be neural activity signals from individual cortical neurons recorded by implanted electrodes or cortical neural ensembles recorded by EEG electrodes. The neural null space is best illustrated by the implementation of brain-machine interfaces (BMIs) *(28)*. Indeed, foundational work shows that both non-human and human primates can learn to control the firing rate of neuronal ensembles or even single neurons in the motor cortex in order to operate a robotic or virtual limb without impairing the control of biological limbs *(29–32)*. BMIs with individual neuron recordings may seem like an ideal candidate for motor augmentation, making it possible to harness dedicated neural patterns for the motor control of extra robotic limbs. However, brain recordings conducted via cortically implanted electrodes are, to date, an unsuitable technique for healthy users, given the severe safety issues that could arise (e.g., surgery-related risks or post-surgery infections). Conversely, the extraction of relevant information by non-invasive BMIs (typically using EEG signals *(28)*) (see example (iii) on Fig. 3B ) is in practice challenging when aimed at controlling XRAs/XRFs. This is because the brain signals of interest would be inevitably mixed with those arising from other cortical activation patterns, including those related to the control of the biological limbs. In unconstrained environments, EEG recordings are also prone to signal contamination due to physiological and non-physiological artefacts (e.g., head or limb motion and power-line interference) *(33)*. In other words, it is difficult to orthogonalize the task null space. Hybrid solutions inspired by collaborative robotics could address the task-intrinsic limitations of human-machine interfaces by employing, for example, intelligent sensorized robotic devices to exploit the shared control paradigm for the low-level kinematic calculations that can extend dexterity in XRAs and XRFs.

## III. Sensory feedback for extra robotic arms and fingers

Numerous studies have demonstrated the importance of somatosensory feedback for dexterous and intuitive motor control in healthy subjects *(34)* and in amputee patients for efficiently controlling a prosthetic device *(2, 35–37)*. Yet, due to the "*Neural Resource Allocation"* problem facing augmentation technology, there are currently only few examples of sensory feedback for XRFs and XRAs (see Table 1). That is, akin to motor control, it is a unique challenge to deliver sensory information about extra robotic limbs without incurring costs to the sensory resources allocated to other body parts. This requires exploiting what we define here as "**sensory complementary space**" (Fig. 5), which consists of all the sensory information that can be provided without interfering with the sensory flow coming from the biological limbs. The sensory feedback can be presented to the limbs involved in the task ("task-intrinsic sensory complementary space") or to another body part ("task-extrinsic sensory complementary space"). The need to identify a null space poses a unique challenge to augmentation that is not faced by other levels of enhancement such as restoration. But the literature on sensorized prostheses for amputees *(38, 39)*, as well as non-invasive approaches inspired by sensory substitution and sensory remapping displays *(35)*, can inspire potential solutions for implementing sensory feedback in extra robotic limbs. For instance, sensory feedback displays for XRAs and XRFs would rely on the user's natural sensory pathways to convey information from artificial receptors to cope with a sensory deficiency – that of the extra robotic limbs – without interfering with other perceptual processes.

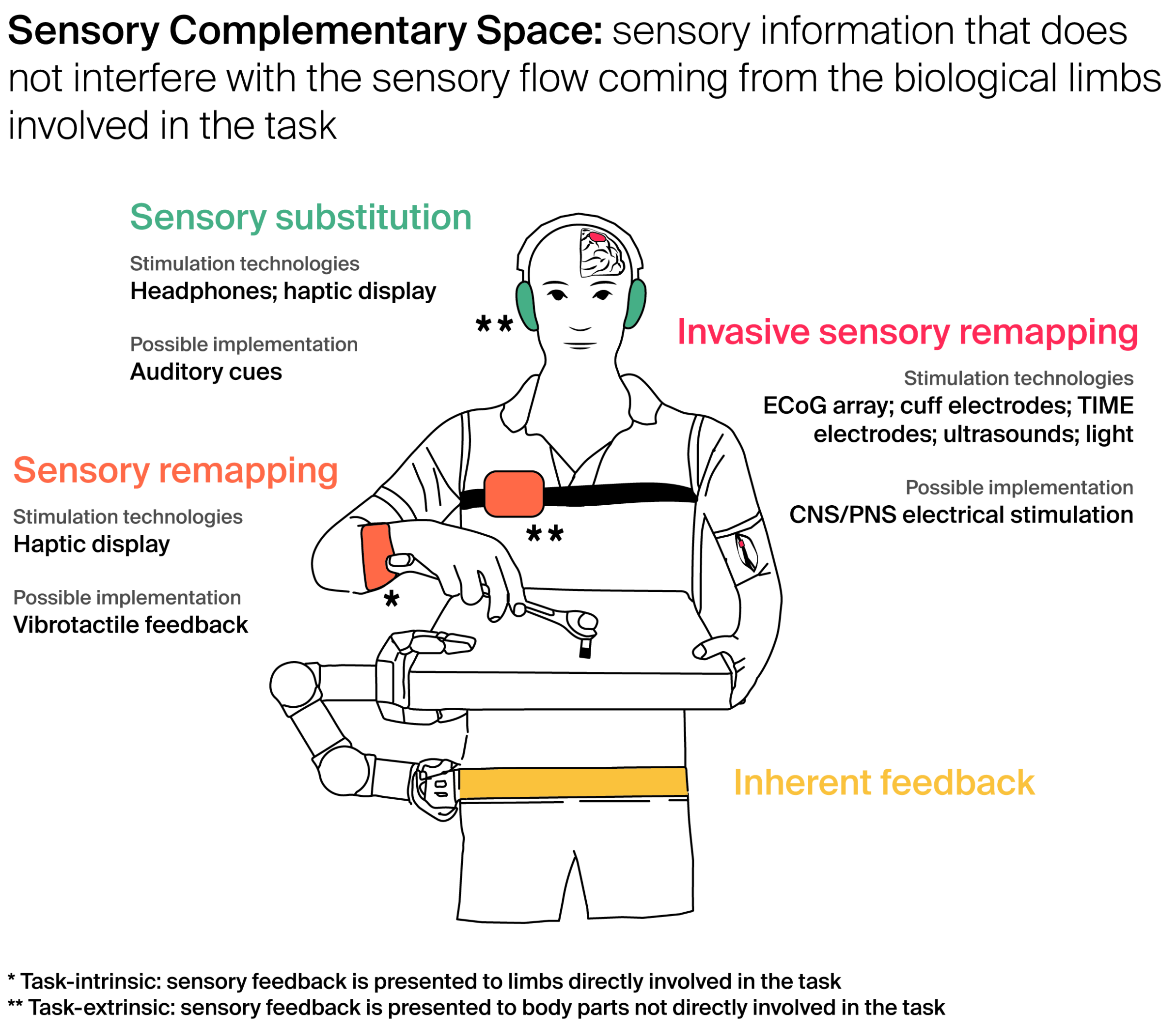

**Figure 5:** Examples of the sensory complementary space via sensory substitution (green) and invasive (red) or non-invasive (orange) sensory remapping. Stimulation technologies and possible implementations are reported for the different approaches. One asterisk denotes task-intrinsic sensory complementary space implementation while two asterisks task-intrinsic ones. The inherent feedback at the interface between the device and the body is also reported (yellow).

Given the wearable nature of extra robotic limbs, it is important to note that a somatosensory input is inherently present at the interface between the device and the user's body, providing implicit sensory feedback. In a series of studies, motor control was found to be improved when physically wearing an XRA, operated via the task-extrinsic muscular null space, compared to controlling a simulated XRA *(12, 40)*. This indicates that there might already be some valuable sensory feedback inherent to wearing an extra robotic limb that supports its motor control. Alternatively, sensory information cues that typically act at different time scales (e.g., event-driven feedback *(36)*), can also be useful for XRA/XRF-body integration. For example, haptic feedback provided via a vibrator worn on the contralateral hand (i.e., task-intrinsic) can improve performance in a hand-robot coordination task *(10)*. The able-bodied participants performed the task with an XRF attached to their wrist, which they controlled using a switch placed on a worn ring (task-intrinsic kinematic null space). With the sensory feedback, a significant reduction in task completion time and mean force exerted was achieved, suggesting that it helps users learn to efficiently control the XRF with less effort and in collaboration with the biological fingers. But what if we managed to precisely target the sensory pathways of the nervous system to report sensory feedback from the extra robotic limb? Akin to what was discussed in the motor control section, it is conceptually feasible to build on existing redundancies in our somatosensory system to convey tactile information. This is demonstrated by the use of invasive somatosensory stimulation *(35, 41, 42)*, showing promising results that are unmet by non-invasive techniques *(37)*. Such an approach was exploited before *(43)*, by "remapping" a sensory pathway in order to provide position feedback to trans-radial amputees. This and related examples *(44)* demonstrate that new associations can be formed and sensory signals used to convey new information. But the analogy to the motor domain ends here because, to our knowledge, stimulation of both peripheral and central nerves in the somatosensory system always induces sensations on the biological body. In other words, while it might be possible to control an extra robotic limb using the motor cortical null space, we do not yet know how to best provide sensations for a non-existing body part. Moreover, it is not clear if users are even able to process additional sensory information in case of increased cognitive load. Preliminary evidence seems to suggest that invasive approaches might be more robust in this case *(37)*. Finally, it is important to emphasize that different types and applications of extra robotic limbs will require different customized solutions for sensory feedback, so a "one size fits all" approach is not suitable.

## IV. Regulatory, legal and ethical considerations

Human augmentation via XRAs and XRFs raises societal and ethical concerns that should be addressed, especially by those collaborating to realize motor augmentation as an industrial or household product (Fig. 1). Some of these challenges are not new and have been debated in other domains, such as industrial automation and digitalization *(45)* or plastic surgery. However, extra robotic limbs increase the complexity of the debate. This is because on one side, there are usage-related risks, ranging from employment considerations to military use. But on the other, there are also philosophical and ethical implications to augmenting the natural physical constraints of a biological body. Recently, an ethical framework has been proposed for human augmentation *(46)*, building on previous frameworks for emerging technologies, such as the Transhumanist Declaration, and providing stakeholders with a starting point for discussion.

A broader and fairer acceptance of augmentation technologies also requires international standards and guidelines aimed at ensuring, among others, safety, equity, equality, and privacy *(47)*. In this respect, a critical step in defining a legal framework for human augmentation was undertaken by the European Union's reform of the regulatory scheme for medical devices, in which the category of "products without an intended medical purpose" was introduced. This includes devices with an augmentative purpose along with analogous therapeutic devices, now better reflecting the enhancement continuum from restoration to augmentation. While this legislation filled a critical legal gap, it mainly deals with safety and security concerns and a unified framework of criteria for permanently or temporarily integrating augmentative technologies with the body is yet to be conceived. From the neuroscience perspective, such criteria are especially important because persistent usage of extra robotic limbs might interfere with a user's biological body representation and potentially even cause disruptions that impair them in daily life *(13)*. If this is the case, humans may find themselves exposed to new vulnerabilities caused by the very technologies designed to overcome their frailty *(48, 49)*. Since the sensorimotor body representation does not fully mature before the age of around 12 *(50)*, children and adolescents could be particularly vulnerable to the impact of these technologies. In sum, the benefits of the availability of augmentation technologies would need to be evaluated against the backdrop of measures protecting its users in general and minors in particular.

## V. Further considerations for successful implementation of motor augmentation

Despite promising results from clinical, neuroscience and engineering research *(13, 17, 20)*, the field of upper-limb augmentation is still in its infancy and many challenges lie ahead. We consider the need to understand how to effectively implement sensorimotor control of these devices without interfering with the biological body a crucial first challenge. Here, we offered the concepts of "motor task null space" and "sensory complementary space" to help focus the initial discourse and pave the way for innovative solutions to the *"Neural Resource Allocation"* problem. Notably, each solution has its own advantages and challenges, which will need to be evaluated for the specific usage scenarios of a given technology. As devices evolve for more diverse functions and settings, these issues will become more complex and require more sophisticated solutions (e.g., invasive implants), which should be accompanied by further ethical and legal oversight.

From a practical perspective, if in order to control an extra robotic limb the user has to limit what they can do with their biological limbs, then there is no enhancement of capabilities but rather a substitution. At the early stages of engineering research, such considerations are usually disregarded. However, without clear criteria for assessing XRAs and XRFs as augmentative devices, the formation of a common ground for reproducible research and progress in the field is limited (see Table 1 for the various different outcome measures used). A unified framework is necessary to evaluate any XRA or XRF in terms of the functional improvements it achieves for the specific task it has been designed for and its versatility for more general use. Relatedly, it is important to consider the extent of training needed to proficiently operate the device – will the effort be comparable to learning to cycle, play the drums or use a new touch screen? Device control that requires a lot of mental resources, such as attention, might introduce a cost to augmentation that users are not able or willing to pay.

Finally, we need to ask critically whether the reliance on motor and sensory pathways requires the XRA/XRF to be represented in the brain as a part of our biological body? Current evidence is suggesting that the neural body representation may not be malleable enough to integrate extra robotic limbs along with the biological limbs *(51, 52)*. As such, instead of aiming to integrate the XRF/XRAs into an already existing body model, our brains might develop new functionally dedicated representations for the extra robotic limbs. This idea relies on Hebbian experience-dependent plasticity mechanisms and the notion that there are redundancies in our nervous system. By freeing the design of extra robotic limbs from the constraints of the body model, we can aim to achieve what we call 'soft embodiment'. This does not entail abandoning the resources that the brain has already evolved, but rather recycling them in a new and better way by taking advantage of the opportunities that non-biological materials and machinery offer *(6)*. Unlike widespread notions of embodiment, soft embodiment is more liberal and does not require the extra robotic limb to be viewed, perceived or experienced as a biological body part. Ultimately, progress for the emerging field of human body augmentation will not only rely on uniting perspectives and concerns from engineering and neuroscience to create and evaluate better devices, but also more widely on legal and ethical considerations.

## Acknowledgments

GD was funded by the European Research Council (ERC) under the European Union's Horizon 2020 research and innovation programme (Grant agreement No. 813713). SS was funded by the CHRONOS project, the Wyss Center for Bio and Neuroengineering, and the Bertarelli Foundation. FV was funded by the ANR grants ANR-17-EURE-0017 FrontCog and ANR-16-CE28-0015 Developmental tool. TRM was funded by a ERC Starting Grant (715022 EmbodiedTech) and a Wellcome Trust Senior Research Fellowship (215575/Z/19/Z). GS and DP were supported by Progetto Prin 2017 "TIGHT: Tactile InteGration for Humans and arTificial systems", prot. 2017SB48FP. SM was funded by the Swiss National Science Foundation through the National Centre of Competence in Research (NCCR) Robotics, the European Research Council (ERC) under the European Union's Horizon 2020 research and innovation programme (Grant agreement No. 813713) and the Bertarelli Foundation.